\documentclass[a4paper, 10pt]{article}      

\usepackage[square,numbers,sort&compress]{natbib} 
\usepackage[pdftex]{graphicx}
\usepackage{amsmath} 
\usepackage{amssymb}  
\usepackage{url}

\title{
Highly comparative fetal heart rate analysis
}
\date{}

\author{B. D. Fulcher\footnote{B. D. Fulcher is with the Department of Physics, University of Oxford, UK:
        {\tt\small ben.d.fulcher$@$gmail.com}}
        , A. E. Georgieva\thanks{A. Georgieva is with the Nuffield Department of Obstetrics and Gynaecology and with the Institute of Biomedical Engineering, University of Oxford: {\tt\small antoniya.georgieva$@$obs-gyn.ox.ac.uk}},
        C. W. G. Redman\thanks{C.\,W.\,G. Redman is with the Nuffield Department of Obstetrics and Gynaecology and also with the Oxford Biomedical Research Centre, University of Oxford: {\tt\small christopher.redman$@$obs-gyn.ox.ac.uk}},
        and Nick S. Jones\thanks{Nick S. Jones is with the Department of Physics, University of Oxford, UK and also with the Department of Mathematics, Imperial College, London: {\tt\small nick.jones$@$imperial.ac.uk}}
}

\begin{document}

\maketitle
\thispagestyle{empty}
\pagestyle{empty}

\begin{abstract}
	A database of fetal heart rate (FHR) time series measured from 7\,221 patients during labor is analyzed with the aim of learning the types of features of these recordings that are informative of low cord pH.
	Our `highly comparative' analysis involves extracting over 9\,000 time-series analysis features from each FHR time series, including measures of autocorrelation, entropy, distribution, and various model fits.
	This diverse collection of features was developed in previous work \cite{Fulcher13}, and is publicly available.
	We describe five features that most accurately classify a balanced training set of 59 `low pH' and 59 `normal pH' FHR recordings.
	We then describe five of the features with the strongest linear correlation to cord pH across the full dataset of FHR time series.
	The features identified in this work may be used as part of a system for guiding intervention during labor in future.
	This work successfully demonstrates the utility of comparing across a large, interdisciplinary literature on time-series analysis to automatically contribute new scientific results for specific biomedical signal processing challenges.
\end{abstract}

\section{Introduction}

During birth, a baby's oxygen supply can be compromised and cause birth asphyxia (suffocation).
Birth asphyxia can lead to seizures, permanent brain damage, and the death of the newborn.
Intervention in the form of a Caesarean section, or the use of forceps or ventouse (vacuum), is required to prevent this chain of events, but such interventions can themselves cause complications and would preferably be avoided.
Currently, the decision to intervene is made on the basis of an electronic recording of the baby's heart rate during labor, a cardiotocogram (CTG).
The mechanisms underlying this recording are complex and its analysis by eye is highly unreliable, whereby different experts can make conflicting decisions on the basis of the same CTG trace \cite{Westgate09}.
This subjectivity in decision-making can also lead to litigation when an `incorrect' decision results in a complication.
These factors have led to a push for research into an objective, computerized system for analyzing CTG recordings to assist the decision-making process \cite{Westgate09}.
Previous reports on this area have been plagued by very small datasets (typically containing less than 500 time series) \cite{Pello91, Warrick10, Chudacek11}; it is difficult to reach reliable conclusions using such datasets for which so few compromised cases are available. 
The present work is distinguished both by the large size of the dataset: 7\,221 FHR time series, and the scale of the analysis: over 9\,000 time-series analysis features are compared.

Our primary aim in this paper is to contribute to a system being developed for intrapartum CTG analysis, {\it OxSys} \cite{Georgieva11}, by providing a set of useful features derived from FHR time series.
Rather than devising new types of features or manually comparing a small number of hand-picked candidates, we take the somewhat unusual approach in this work of comparing simultaneously the performance of thousands of features developed across the scientific time-series analysis literature.
Using this highly comparative approach, those features that are the most successful are retrieved and subsequently analyzed and interpreted.

\section{Data and Methods}

\subsection{Data} \label{sec:data}
\begin{figure}[h]
	\centering
		\includegraphics[width = 11cm]{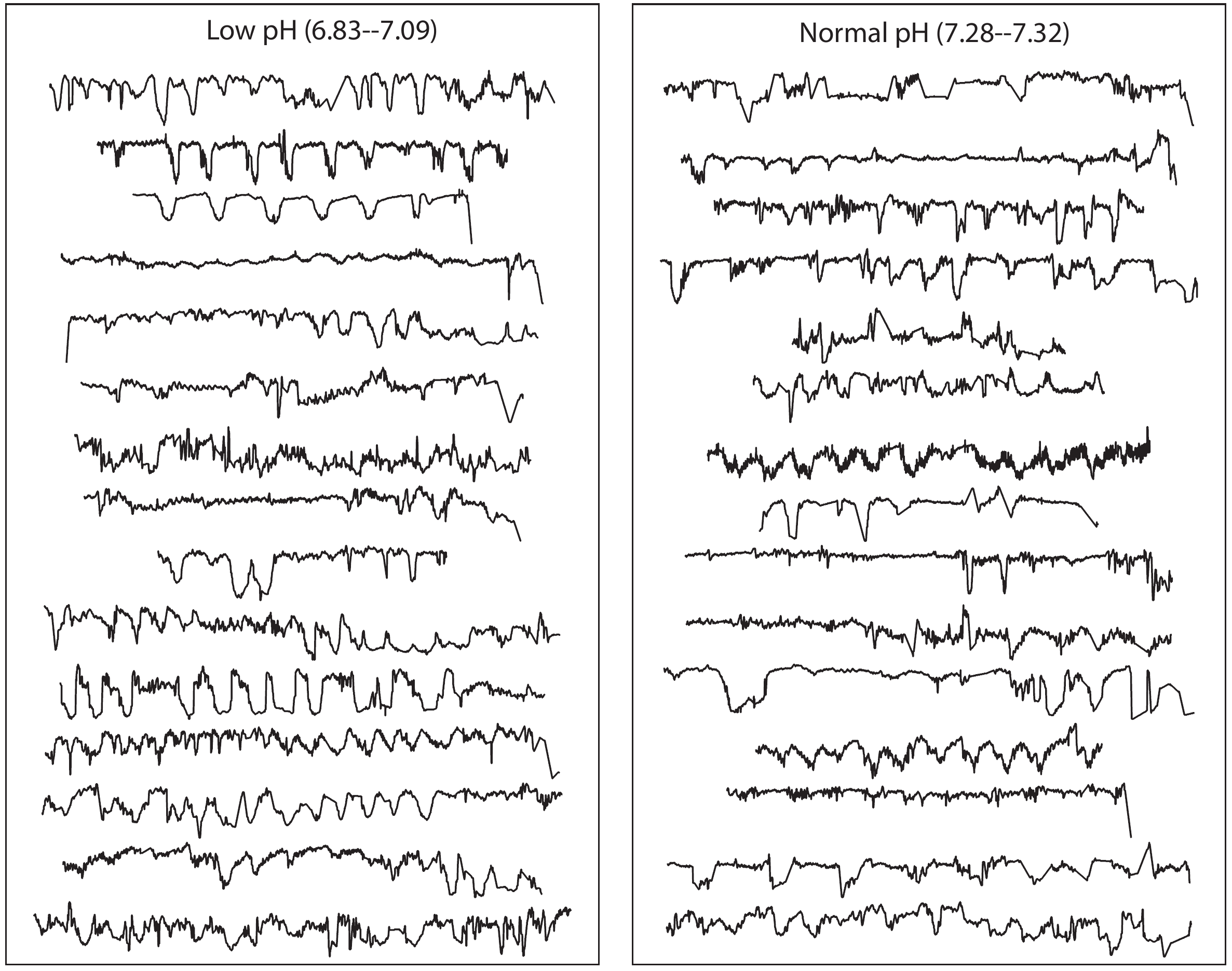}
	\caption{
	\textbf{Fetal heart rate time series in each of two classes: {\it\bf low pH} and {\it\bf normal pH}}.
	The plotted time series are from the training set and span the full range of pH values in each group, which is given in parentheses.
	}
	\label{fig:training_timeseries}
\end{figure}

The initial dataset analyzed in this paper contains 7\,568 FHR time series sampled at 4\,Hz and recorded in the last 30\,min before delivery.
The data met a set of quality-based criteria from an initial set of 107\,614 deliveries in John Radcliffe hospital, Oxford, UK between 20 April 1993 and 28 February 2008 \cite{Georgieva11}, and were preprocessed to remove various known artifacts \cite{Georgieva10}. 
The data were processed further in this work: by linearly interpolating short durations of missing values, trimming longer durations of missing values, and removing time series with a large proportion of missing values, resulting in a dataset containing 7\,221 FHR recordings.
The data were partitioned into balanced training and test sets according to a previous study \cite{Georgieva11}.
Within each set, each FHR recording is classified according to the cord pH of the corresponding baby, as either {\it low pH} ($\leq 7.1$) or {\it normal pH} ($>7.1$).
The training set contains 59 time series of each class, and the the test set contains 117 time series of each class.
Examples of both classes of time series in the training dataset are shown in Fig.~\ref{fig:training_timeseries}.

\subsection{Highly comparative analysis}

Our highly comparative time-series analysis method is outlined in this section, and is described in detail elsewhere \cite{Fulcher13}.
The method relies on a collection of 9\,613 algorithms for extracting features from time series.
These algorithms span a large variety of time-series properties, summarizing their autocorrelation, stationarity, summaries of their power spectra, wavelet decompositions, their distribution of values, fits to various time-series models (e.g., autoregressive, Gaussian Process, and Hidden Markov models), measures from the physical nonlinear time-series analysis literature (e.g., correlation dimension estimates, nonlinear prediction errors, fractal scaling properties), information theoretic quantities (e.g., permutation entropy, Sample Entropy, Lempel-Ziv Complexity), and others \cite{Fulcher13}.
Each of these myriad methods is encoded in the same way: as an algorithm that maps an input time series to a single real number.
Code for all time-series analysis methods used can be downloaded and explored at \url{www.comp-engine.org/timeseries}.

To compare their performance, all of these features were evaluated on all FHR time series in the dataset.
Some algorithms could not be applied appropriately to some time series, e.g., fitting a positive-only distribution to time series that are not positive-only.
In cases such as these, algorithms returned a {\it special value}: an infinity or a NaN.
Features for which this occurred at least once across the dataset were removed from our analysis, and in this way, the initial set of 9\,613 features was reduced to approximately 7\,600 features.

\subsection{Classification and Clustering}

Classification rates quoted throughout this paper were obtained from a simple linear discriminant classifier, implemented using the \texttt{classify} function from Matlab's {\it Statistics Toolbox}\footnote{We used Matlab 2011a. Matlab is a product of The MathWorks, Natick, MA.}, which provides a highly intuitive and interpretable result: a linear partition of the feature space \cite{Hastie09}.
For the single features focused on in this paper, linear classification boundaries are simply thresholds on the value of each feature.

Clustering is used to automatically reduce sets of features to smaller, representative subsets in this work.
We used average linkage clustering, as implemented using the \texttt{linkage} function from Matlab's {\it Statistics Toolbox}.

\section{Results} \label{sec:results}

\subsection{Classification} \label{sec:FHR_classification}
\begin{figure}[h]
	\centering
		\includegraphics[width = 11cm]{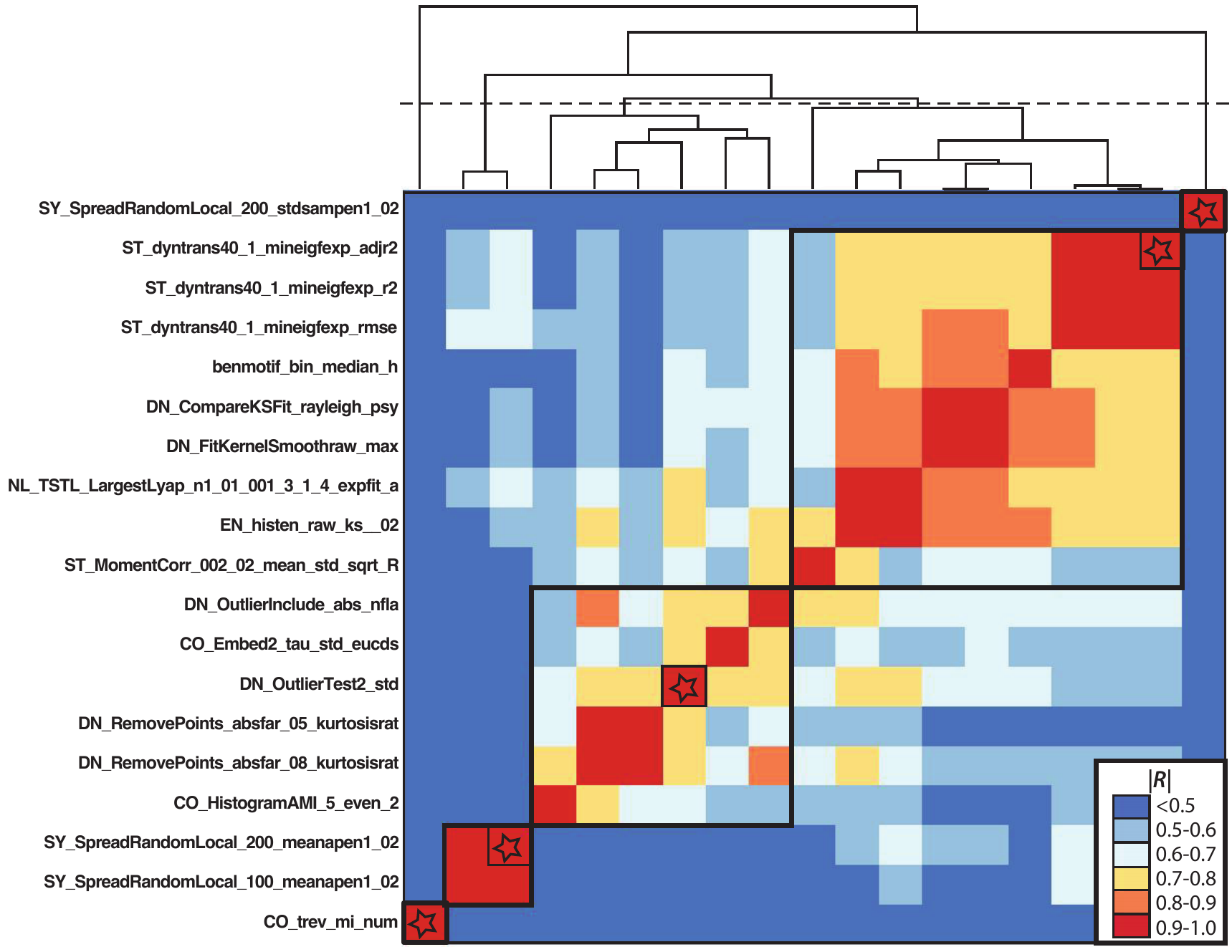}
	\caption{
	\textbf{Clustering is used to select five features that best represent the nineteen features with a misclassification rate under 30\%.}
	The magnitude of linear correlation coefficients, $|R|$, calculated between all pairs of the top nineteen features are plotted as a colored matrix.
	The name of each feature is labeled to the left of the plot.
	A dendrogram constructed using average linkage clustering is plotted above the pairwise correlation matrix, and is cut at the point plotted with a dashed line to create five clusters of features.
	The resulting clusters are represented using black squares in the pairwise correlation matrix.
	The features with the lowest misclassification rates in each cluster are selected to represent that cluster, and are indicated using stars in the correlation matrix.
	The performance of each of these features on the test set is illustrated in Fig.~\ref{fig:training_top5ks}.
	}
	\label{fig:training_19totop5}
\end{figure}

First we analyze the balanced training and test sets described above, with the aim of distinguishing FHR time series measured from fetuses with low cord pH at birth.
We calculated the (in-sample) linear misclassification rates for each of 7\,586 features (those with no special-valued outputs) on the training set.
We then selected the nineteen most successful features: those with a false discovery rate \cite{Hastie09} less than 0.001 (cf. \cite{Fulcher13}), corresponding to a linear misclassification rate under 30\%.
Since some of these nineteen features are highly correlated to one another across the dataset, we proceeded to construct a smaller set of features that minimizes this redundancy.
Linear correlation coefficients calculated between all pairs of these nineteen features across the FHR dataset are shown in Fig.~\ref{fig:training_19totop5}.
A dendrogram relating the features was constructed using average linkage clustering and is shown above the pairwise correlation matrix in Fig.~\ref{fig:training_19totop5}.
By thresholding the dendrogram, the features were clustered into five groups.
Features within each cluster have high linear correlations to one another and can be well-summarized by a single representative member.
These representative features were chosen as those with the lowest misclassification rate in each cluster, and are labeled using stars in Fig.~\ref{fig:training_19totop5}.
In this way, a more manageable set of five relatively independent features was identified that effectively summarizes the most successful time-series analysis algorithms for distinguishing babies with low cord pH from FHR time series recorded during labor.

\begin{figure}[h]
	\centering
		\includegraphics[width = 11cm]{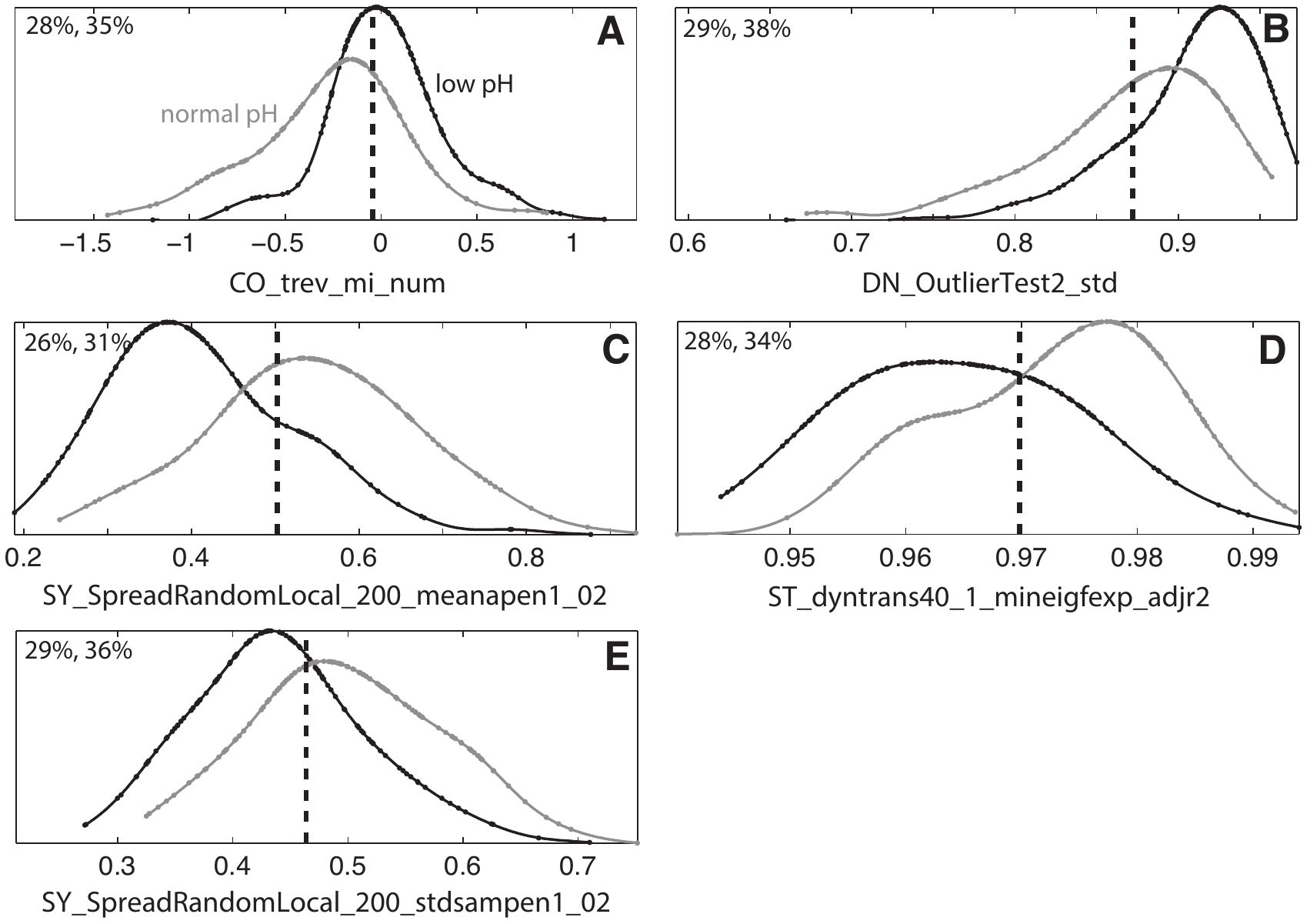}
	\caption{
	\textbf{Five representative features with a misclassification rate under 30\% on the training set show good performance on the test set.}
	The probability distribution for the `low pH' group (black) and the `normal pH' group (gray) are plotted for each feature applied to the balanced test dataset.
	Linear discrimination thresholds learned on the training set are indicated using a dashed black line and are used to classify these testing data.
	Misclassification rates on the training set (the former number) and the test set (the latter number) are annotated to the top left of all plots.
	}
	\label{fig:training_top5ks}
\end{figure}

We now describe these five features, and investigate their performance on the test dataset.
There is insufficient space to describe each feature in detail, but brief summaries are as follows:
(i) \textbf{CO\_trev\_mi\_num} is a quantity related to the time-reversal asymmetry of a time series,
(ii) \textbf{DN\_OutlierTest2\_std} returns the ratio of standard deviations before and after removing 2\% of the highest and lowest values of a time series,
(iii) \textbf{SY\_SpreadRandomLocal\_200\_meanapen1\_02} averages local Approximate Entropy \cite{Pincus91}, ApEn(1,0.2), estimates,
(iv) \textbf{ST\_dyntrans40\_1\_mineigfexp\_adjr2} calculates 1-step transition matrices for different alphabet sizes and fits a decaying exponential to the minimum eigenvalues of these transition matrices, and (v) \textbf{SY\_SpreadRandomLocal\_200\_stdsampen1\_02} measures the variation in local Sample Entropy \cite{Richman00}, \mbox{SampEn(1,0.2)}, estimates from the time series.
Distributions of the outputs of each of these features on the test data are shown in Fig.~\ref{fig:training_top5ks}.
These distributions provide an interpretable difference in the properties of the two groups of FHR time series: e.g., as shown in Fig.~\ref{fig:training_top5ks}B, we see that healthy FHR recordings (gray) typically have more extreme outliers (and hence lower values of \textbf{DN\_OutlierTest2\_std}) compared to the low pH group (black).
As indicated in Fig.~\ref{fig:training_top5ks}, using a simple threshold on the output of each feature, in-sample misclassification rates range from 26\%--29\%, and out-of-sample misclassification rates range from 31\%--38\%.
We note that classifiers that combine multiple features for this dataset (constructed using greedy forward feature selection \cite{Fulcher13}) showed no improvement in out-of-sample performance over single-feature classifiers.

\subsection{Regression onto arterial cord pH} \label{sec:FHR_regression}
We now investigate features with outputs that correlate linearly with the cord pH across the full dataset of 7\,221 FHR time series.
The magnitude of linear correlation coefficients, $|R|$, were low, with $|R| < 0.3$, but significantly larger than would be expected by chance (i.e., when the output of features are shuffled at random, cf. multiple hypothesis testing \cite{Hastie09, Fulcher13}).
In an analogous method to that shown for the classification task above, clustering was used to construct a set of five features that are representative of those with the strongest linear correlation coefficients, $|R|$, to cord pH.
These five features are now described briefly, with correlation coefficients given in parentheses:
(i) \textbf{coeff\_var\_2} ($R = -0.28$) returns the second order coefficient of variation: $(\sigma/\mu)^2$, where $\sigma$ and $\mu$ are the standard deviation and mean of the time series, respectively,
(ii) \textbf{median\_absolute\_deviation} ($R = -0.28$) returns the median absolute deviation, $\langle |x-\mathrm{median}(x)| \rangle$, a measure of spread of the time series, $x$,
(iii) \textbf{ST\_dyntrans40\_1\_mineigfexp\_adjr2} ($R = 0.25$) is a quantity derived from transition matrices, as described for the classification task above,
(iv) \textbf{DN\_SimpleFit\_exp1\_rmse\_h30} ($R = 0.25$) returns the goodness of an exponential fit to the distribution of time-series values, and
(v) \textbf{CO\_Embed2\_tau\_arearat} ($R = -0.24$) returns the ratio of areas spanned by points in a two-dimensional time-delay embedding space for the time series \cite{Kantz04}.
Interpreting the sign of the correlation also allows us to interpret the results directly; for example, \textbf{DN\_SimpleFit\_exp1\_rmse\_h30} has $R = 0.25$, revealing that FHR recordings associated with a higher cord pH have distributions that are typically closer to exponential than those with lower cord pH.



\subsection{EveREst plots} \label{sec:FHR_combined_set}

The great majority of FHR recordings studied in this work correspond to healthy babies with normal cord pH.
For example, consider the following three groups: (i) {\it low pH}, defined as patients with an arterial cord pH $\leq 7.05$, contains 302 patients, (ii) {\it compromised}, defined as patients with a reported severe, moderate, or mild reported compromise \cite{Georgieva11}, contains 795 patients, and (iii) {\it low pH and compromised}, defined as patients that fulfill both of the above criteria, contains just 110 patients.
These problematic cases may be preventable and are the most interesting to clinicians who must decide whether an intervention is appropriate in real time during labor.
Distinguishing such small numbers of problematic scenarios from a large total cohort of 7\,221 patients is difficult.
One way of proceeding, which we follow here, is to divide the total cohort into $N_\mathrm{group}$ equally-populated groups and compare the proportion of compromised cases in each group.
A graphical representation of this approach has been termed an {\it Event Rate Estimate} (EveREst) plot \cite{Georgieva11}.

\begin{figure}
	\centering
		\includegraphics[width = 8cm]{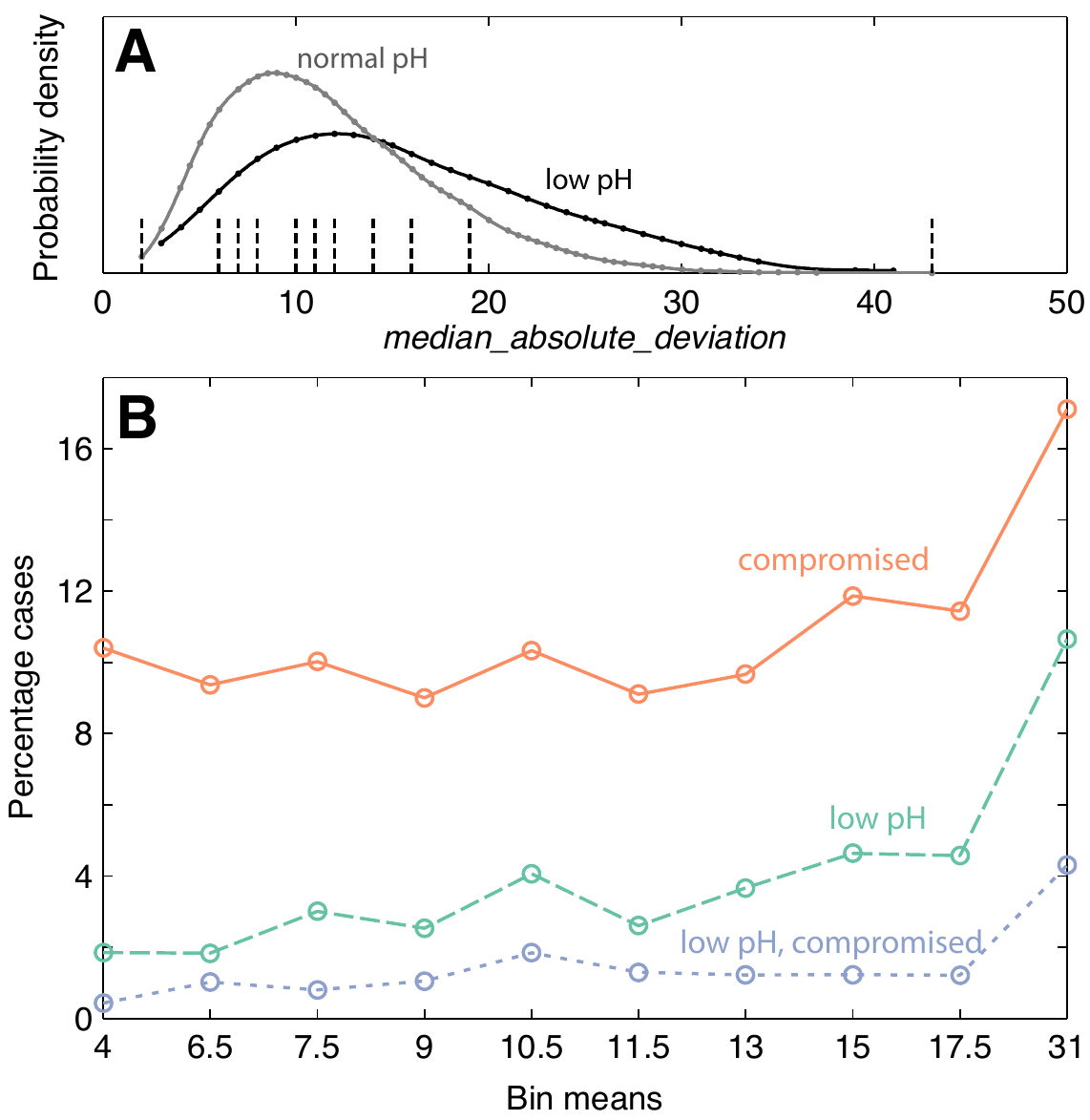}
	\caption{\textbf{Distributions and EveREst plot for a measure of spread feature: median\_absolute\_deviation.}
	\textbf{A} Distributions are plotted of the {\it low pH} (black) and {\it normal pH} (gray) groups defined by a pH threshold of 7.05.
	There are 302 FHR recordings with a corresponding arterial cord pH $\leq 7.05$, and 6\,919 with pH $ > 7.05$.
	\textbf{B} The EverEst plot was generated by dividing the 7\,221 patients into 10 equally-populated groups, ordered by their {\it median\_absolute\_deviation}, $\langle |x-\mathrm{median}(x)| \rangle$.
	The partitions that define these equally-populated groups are shown as dashed lines in the upper plot; each group is represented by its mean in the EverEst plot.
	Three types of compromised patients are represented in the EveREst plot: (i) low pH (dashed green line, and represented as distributions in \textbf{A}), (ii) compromised (orange line), and (iii) both low pH and compromised (dotted blue line).
	A useful predictor of compromised babies would involve simply measuring the {\it median\_absolute\_deviation} of FHR time series during labor: a value in the highest decile indicates an increased risk of low pH and compromise.
	}
	\label{fig:mead_everest}
\end{figure}

By ordering all FHR time series according to the value of a given feature, we constructed EveREst plots using $N_\mathrm{group} = 10$ for the successful features selected above.
An example is shown in Fig.~\ref{fig:mead_everest} for the {\it median\_absolute\_deviation} measure of spread, $\langle |x-\mathrm{median}(x)|\rangle$, which was selected from the regression task described above.
Compared to the distributions shown in Fig.~\ref{fig:training_top5ks} for a balanced dataset, the distribution in Fig.~\ref{fig:mead_everest}A requires a more subtle interpretation, as the low pH condition (plotted black) contains just 302 recordings, compared to the 6\,919 recordings with normal pH (plotted gray).
However, dividing the patients into equal groups, as in Fig.~\ref{fig:mead_everest}B, reveals the proportion of problematic patients in each equally-populated group, which can be used to determine thresholds by which the two groups could be separated.
Note that other features selected in the classification and regression tasks above have qualitatively similar EveREst plots to that shown in Fig.~\ref{fig:mead_everest}B.

For this {\it median\_absolute\_deviation} feature, there is a relatively sharp rise in low pH and compromised cases in the final bin.
Patients in this bin: with $\langle |x-\mathrm{median}(x)| \rangle > 19.25$, therefore have an increased risk of both delivering a baby with compromise, and of delivering a baby with low cord pH.
As with most real-world applications, predicting compromise or low cord pH from FHR recordings is an complex and subtle problem that depends on a large number of variables.
Thus, although not clinically useful on its own, this simple {\it median\_absolute\_deviation} measure of spread is both informative and extremely easy to compute.
As much as an eight-fold increase in risk is observed here (for the low pH and compromised group in the final bin), making this feature a good candidate for further investigation in future work.

\section{Conclusions} \label{sec:FHR_conclusions}

Five representative features were selected from those that were most successful at classifying FHR time series, and another five were selected to represent those with the strongest linear correlations to arterial cord pH across a dataset of 7\,221 FHR time series.
One of these features occurs in both sets, and hence we have a resulting set of nine candidate features that will be investigated as part of the Oxford System for intrapartum CTG analysis: {\it OxSys} \cite{Georgieva11}.
Combined with features from other CTG recordings and additional clinical data, future work will focus on using these features to build a commercial diagnostic system---an intrapartum analogue of the established Dawes-Redman system \cite{Pardey02}.

Using the example of FHR analysis, this paper demonstrates the broad applicability of our highly comparative time-series analysis methodology \cite{Fulcher13}.
The empirical structure of the labeled data was used to select features automatically, from a diverse and interdisciplinary scientific literature on time-series analysis.
Although our set of over 9\,000 features is far from exhaustive, we have successfully identified some of the most promising features from what is a comprehensive collection, and shown how they can be interpreted in the context of this FHR analysis problem.
Extensive further work will be required to interpret the new features clinically, to integrate them into {\it OxSys}, and to study their relationships with existing FHR features.
These candidate features will ultimately become components of multivariate analyses including other FHR features: standard morphological features (e.g., baseline, deceleration, variability) and clinical information about labour (e.g., use of epidural, gestation age).
Code for all time-series analysis methods used here can be explored and downloaded at \url{www.comp-engine.org/timeseries}.

\addtolength{\textheight}{-12cm}   



%




\bibliographystyle{benthesisnum}


\end{document}